\documentclass[letterpaper]{article} 
\usepackage{aaai24}  
\usepackage{times}  
\usepackage{helvet}  
\usepackage{courier}  
\usepackage[hyphens]{url}  
\usepackage{graphicx} 
\usepackage{natbib}  
\usepackage{caption} 
\usepackage{algorithm}
\usepackage{algorithmic}
\usepackage{lineno}
\usepackage[dvipsnames]{xcolor}
\usepackage{amsmath}
\usepackage{cite}
\usepackage{newfloat}
\usepackage{listings}
\usepackage{booktabs}
\usepackage{makecell}
\usepackage{multirow}
\usepackage{eso-pic}
\usepackage{times}
\usepackage{epsfig}
\usepackage{graphicx}
\usepackage{amsmath}
\usepackage{amssymb}
\usepackage{xcolor}
\usepackage{amssymb}

\definecolor{nbarrier}{RGB}{255, 120, 50}
\definecolor{nbicycle}{RGB}{255, 192, 203}
\definecolor{nbus}{RGB}{255, 255, 0}
\definecolor{ncar}{RGB}{0, 150, 245}
\definecolor{nconstruct}{RGB}{0, 255, 255}
\definecolor{nmotor}{RGB}{200, 180, 0}
\definecolor{npedestrian}{RGB}{255, 0, 0}
\definecolor{ntraffic}{RGB}{255, 240, 150}
\definecolor{ntrailer}{RGB}{135, 60, 0}
\definecolor{ntruck}{RGB}{160, 32, 240}
\definecolor{ndriveable}{RGB}{255, 0, 255}
\definecolor{nothers}{RGB}{0, 0, 0}
\definecolor{nother}{RGB}{139, 137, 137} 
\definecolor{nsidewalk}{RGB}{75, 0, 75}
\definecolor{nterrain}{RGB}{150, 240, 80}
\definecolor{nmanmade}{RGB}{230, 230, 250}
\definecolor{nvegetation}{RGB}{0, 175, 0}

\DeclareCaptionStyle{ruled}{labelfont=normalfont,labelsep=colon,strut=off} 
\floatstyle{ruled}
\newfloat{listing}{tb}{lst}{}
\floatname{listing}{Listing}
\title{RadOcc: Learning Cross-Modality Occupancy Knowledge through  \\ Rendering Assisted Distillation}
\author{
Haiming Zhang\textsuperscript{\rm 1,2}\footnote{Work done during an internship at Huawei Noah's Ark Lab.},
Xu Yan\textsuperscript{\rm3\dag},
Dongfeng Bai\textsuperscript{\rm 3},
Jiantao Gao\textsuperscript{\rm 3},
Pan Wang\textsuperscript{\rm 3},\\
Bingbing Liu\textsuperscript{\rm 3},
Shuguang Cui\textsuperscript{\rm2,1},
Zhen Li\textsuperscript{\rm2,1}\footnote{Corresponding authors: Xu Yan and Zhen Li.}
}
\affiliations{
\textsuperscript{\rm 1}FNii, CUHK-Shenzhen, \textsuperscript{\rm 2}SSE, CUHK-Shenzhen, \textsuperscript{\rm 3}Huawei Noah's Ark Lab\\

\texttt{\small\{haimingzhang@link., xuyan1@link., lizhen@\}cuhk.edu.cn}
}

\usepackage{bibentry}

\begin{document}

\maketitle
\begin{abstract}
	
	3D occupancy prediction is an emerging task that aims to estimate the occupancy states and semantics of 3D scenes using multi-view images.
	However, image-based scene perception encounters significant challenges in achieving accurate prediction due to the absence of geometric priors.
	In this paper, we address this issue by exploring cross-modal knowledge distillation in this task, \textit{i.e.,} we leverage a stronger multi-modal model to guide the visual model during training. 
	In practice, we observe that directly applying features or logits alignment, proposed and widely used in bird's-eye-view (BEV) perception, does not yield satisfactory results.
	To overcome this problem, we introduce \textbf{RadOcc}, a \textbf{R}endering \textbf{a}ssisted \textbf{d}istillation paradigm for 3D \textbf{Occ}upancy prediction. 
	By employing differentiable volume rendering, we generate depth and semantic maps in perspective views and propose two novel consistency criteria between the rendered outputs of teacher and student models.
	Specifically, the depth consistency loss aligns the termination distributions of the rendered rays, while the semantic consistency loss mimics the intra-segment similarity guided by vision foundation models (VLMs). 
	Experimental results on the nuScenes dataset demonstrate the effectiveness of our proposed method in improving various 3D occupancy prediction approaches, \textit{e.g.,} our proposed methodology enhances our baseline by {\textbf{2.2\%}} in the metric of mIoU and achieves {\textbf{50\%}} in Occ3D benchmark. 
	
\end{abstract}

\section{Introduction}

3D occupancy prediction (3D-OP) is a crucial task within the field of 3D scene understanding, which has garnered considerable attention, particularly in the field of autonomous driving~\cite{wang2023panoocc,tong2023scene,tian2023occ3d}.
In contrast to other 3D perception tasks, such as object detection using bounding box representations, 3D-OP involves the simultaneous estimation of both the occupancy state and semantics in the 3D space using multi-view images~\cite{tian2023occ3d}. 
This is achieved by leveraging geometry-aware cubes to represent a wide range of objects and background shapes.

\begin{figure}[t]
	\centering
	\includegraphics[width=\linewidth]{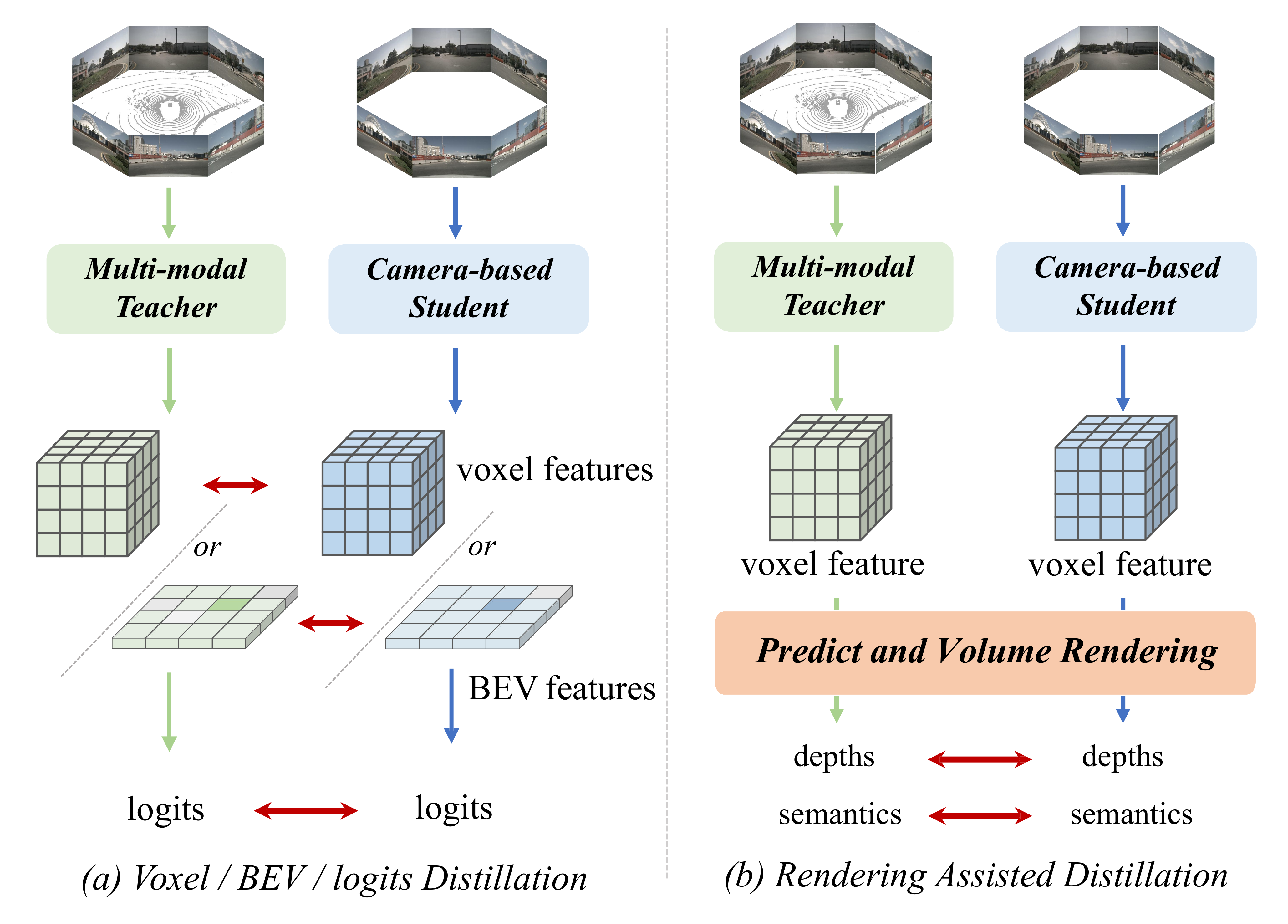} 
	
	\caption{\textbf{Rendering Assisted Distillation}. (a) Existing methods conduct alignment on features or logits. (b) Our proposed RadOcc method constrains the rendered depth maps and semantics simultaneously.}
	\label{fig1:comparison}
\end{figure}

In the realm of 3D occupancy prediction, remarkable advancements have been achieved thus far. 
These advancements have been made possible by adopting a pipeline inspired by Bird's Eye View (BEV) perception, which utilizes either forward projection~\cite{huang2021bevdet} or backward projection~\cite{li2022bevformer} techniques for view transformation.
This process generates 3D volume features that capture the spatial information of the scene, which are then fed into the prediction head for occupancy predictions.
However, relying solely on camera modality poses challenges in accurate prediction due to the lack of geometric perception.
To overcome this bottleneck, two mainstream solutions have emerged in the field of BEV perception:
\textbf{1)} integrating geometric-aware LiDAR input and fusing the complementary information of the two modalities~\cite{liu2023bevfusion}, and
\textbf{2)} conducting knowledge distillation to transfer the complementary knowledge from other modalities to a single-modality model~\cite{zhou2023unidistill}.
As the first solution introduces additional network designs and computational overhead, recent works have increasingly focused on the second solution, aiming to develop stronger single-modal models through distilling multi-modal knowledge.

In this paper, we present the first investigation into cross-modal knowledge distillation for the task of 3D occupancy prediction. 
Building upon existing methods in the field of BEV perception that leverage BEV or logits consistency for knowledge transfer, we extend these distillation techniques to aligning voxel features and voxel logits in the task of 3D occupancy prediction, as depicted in Figure~\ref{fig1:comparison}(a). 
However, our preliminary experiments reveal that these alignment techniques face significant challenges in achieving satisfactory results in the task of 3D-OP, particularly the former approach introduces negative transfer.
This challenge may stem from the fundamental disparity between 3D object detection and occupancy prediction, where the latter is a more fine-grained perception task that requires capturing geometric details as well as background objects.

To address the aforementioned challenges, we propose \textbf{RadOcc}, a novel approach that leverages differentiable volume rendering for cross-modal knowledge distillation. 
The key idea of RadOcc is conducting alignment between rendered results generated by teacher and student models, as Figure~\ref{fig1:comparison}(b).
Specifically, we employ volume rendering~\cite{mildenhall2021nerf} on voxel features using the camera's intrinsic and extrinsic parameters, which enables us to obtain corresponding depth maps and semantic maps from different viewpoints. 
To achieve better alignment between the rendered outputs, we introduce the novel Rendered Depth Consistency (\textbf{RDC}) and Rendered Semantic Consistency (\textbf{RSC}) losses. 
On the one hand, the RDC loss enforces consistency of ray distribution, which enables the student model to capture the underlying structure of the data. 
On the other hand, the RSC loss capitalizes on the strengths of vision foundation models~\cite{kirillov2023segment}, and leverages pre-extracted segments to conduct an affinity distillation. 
This criterion allows the model to learn and compare semantic representations of different image regions, enhancing its ability to capture fine-grained details.
By combining the above constraints, our proposed method effectively harnesses the cross-modal knowledge distillation, leading to improved performance and better optimization for the student model.
We demonstrate the effectiveness of our approach on both dense and sparse occupancy prediction and achieve state-of-the-art results on both tasks.

In summary, our main contributions are threefold:

\begin{itemize}
	\setlength{\itemsep}{0pt}
	\setlength{\parsep}{0pt}
	\setlength{\parskip}{0pt}
	\item We propose a rendering assisted distillation paradigm for 3D occupancy prediction, named \textbf{RadOcc}. Our paper is the first to explore cross-modality knowledge distillation in 3D-OP and provides valuable insights into the application of existing BEV distillation techniques for this task.
	\item Two novel distillation constraints, \textit{i.e.,} rendered depth and semantic consistency (\textbf{RDC \& RSC}), are proposed, which effectively enhance the knowledge transfer process through aligning ray distribution and affinity matrices guided by vision foundation models.
	\item Equipped with the proposed methodology, RadOcc achieves state-of-the-art performance on the Occ3D and nuScenes benchmarks for dense and sparse occupancy prediction. Furthermore, we verify that our proposed distillation approach can effectively boost the performance of several baseline models.
\end{itemize}

\section{Related Work}
\subsection{Camera-based 3D Perception}
Camera-based 3D perception has emerged as a significant research focus in the field of autonomous driving, owing to its cost-effectiveness and rich visual attributes.
Recent advancements have aimed to integrate multiple tasks into a unified framework by transforming image-based features into a Bird's Eye View (BEV) space.
One mainstream follows the forward projection paradigm proposed in LSS~\cite{philion2020lift}, where multi-view image features are projected onto the BEV plane through predicted depth maps~\cite{huang2021bevdet,li2023bevdepth,li2022bevstereo}. 
Another mainstream (\textit{i.e.,} backward projection) draws inspiration from DETR3D~\cite{wang2022detr3d}, which involves using learnable queries and a cross-attention mechanism to extract information from image features~\cite{li2022bevformer,lu2022learning,jiang2023polarformer}.
Although these methods effectively compress information onto the BEV plane, they may lose some of the essential structural details inherent in 3D space. 
Introducing LiDAR priors through cross-modal knowledge distillation makes them ideal for understanding the structure of 3D scenes while keeping efficiency. 

\subsection{3D Occupancy Prediction}
The field of 3D occupancy prediction (3D-OP) has garnered significant attention in recent years, with the aim of reconstructing the 3D volumetric scene structure from multi-view images. 
This area can be broadly classified into two categories based on the type of supervision: \textit{sparse prediction} and \textit{dense prediction}. 
On the one hand, sparse prediction methods utilize LiDAR points as supervision and are evaluated on LiDAR semantic segmentation benchmarks. 
For instance, TPVFormer~\cite{huang2023tri} proposes a tri-perspective view method for predicting 3D occupancy, while PanoOcc~\cite{wang2023panoocc} unifies the task of occupancy prediction with panoptic segmentation in a coarse-to-fine scheme. 
On the other hand, dense prediction methods are more akin to the Semantic Scene Completion (SSC) task~\cite{song2017semantic,yan2021sparse}, with the core difference being whether to consider the area that the camera cannot capture.
Recently, several studies focus on the task of dense occupancy prediction and introduce new benchmarks using nuScenes dataset~\cite{caesar2020nuscenes} at the same period, such as OpenOccupancy~\cite{wang2023openoccupancy}, OpenOcc~\cite{tong2023scene}, SurroundOcc~\cite{wei2023surroundocc} and Occ3D~\cite{tian2023occ3d}.
These works mainly adopt the architecture from BEV perception and use 3D convolution to construct an extra head for occupancy prediction.
We find out that concurrent work~\cite{gan2023simple} also utilizes volume rendering technique, however, they naively apply rendered results as auxiliary supervision.
Still, we first time investigate cross-modal knowledge distillation in this field, and our proposed method can be integrated into arbitrary previous works.

\begin{figure*}[t]
	\begin{center}
		\includegraphics[width=\linewidth]{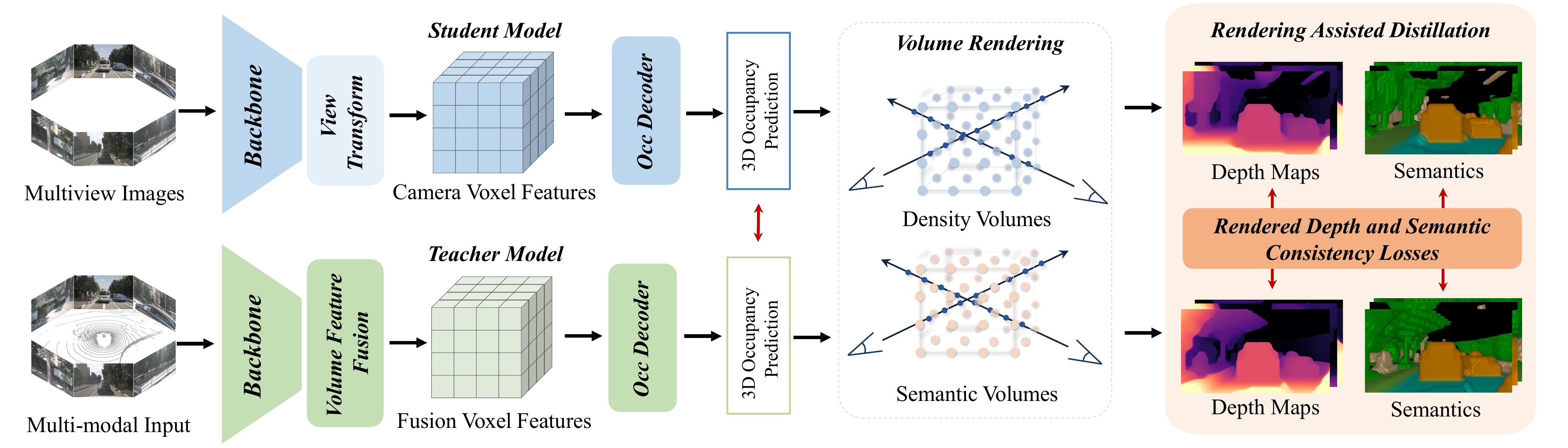}
	\end{center}

	\caption{\textbf{Overall framework of RadOcc.} It adopts a teacher-student architecture, where the teacher network is a multi-modal model while the student network only takes camera inputs. The predictions of two networks will be utilized to generate rendered depth and semantics through differentiable volume rendering. The newly proposed rendered depth and semantic consistency losses are adopted between the rendered results.}
	\label{fig2:framework}

\end{figure*}

\subsection{Cross-Modal Knowledge Distillation} 
Knowledge distillation has been a popular technique in the field of computer vision since its introduction in~\cite{hinton2015distilling}. This technique initially involves compressing a large network (teacher) into a more compact and efficient one (student), while simultaneously improving the performance of the student. Over the years, the effectiveness of knowledge distillation has led to its widespread exploration in various computer vision tasks, including object detection~\cite{dai2021general,guo2021distilling,zhang2020improve}, semantic segmentation~\cite{hou2020inter,liu2019structured} and other tasks~\cite{yan2022let,zhao2023cpu,yuan2022x,zhou2023bev}. 
Recently, knowledge distillation has been introduced into 3D perception tasks for knowledge transfer between models using different modalities. For instance, \cite{chong2022monodistill} transfers depth knowledge of LiDAR points to a camera-based student detector by training another camera-based teacher with LiDAR projected to perspective view. 
2DPASS~\cite{yan20222dpass} utilizes multiscale fusion-to-single knowledge distillation to enhance the LiDAR model with image priors.
In the field of BEV perception, CMKD~\cite{hong2022cross}, BEVDistill~\cite{chen2022bevdistill} and UniDistill~\cite{zhou2023unidistill} perform cross-modality distillation in BEV space. Specifically, these methods transform prior knowledge through distillation in feature, relation, and output levels. Although these efforts have greatly enhanced the performance of student models, they cannot achieve satisfactory performance gains when directly applied to the task of 3D occupancy prediction.

\section{Methodology}
\subsection{Problem Setup}
3D occupancy prediction leverages multiview images as input to predict a semantic volume surrounding the ego-vehicle. 
Specifically, it takes into account the current multiview images denoted as $\mathcal{I}_t = \{\mathcal{I}^t_1,...,\mathcal{I}^t_n\}$, as well as the previous frames $\mathcal{I}^{t-1},...,\mathcal{I}^{t-k}$, where $k$ represents the number of history frames and $n$ denotes the camera view index.
By incorporating this temporal information, the model finally predicts the semantic voxel volume $\mathcal{Y}_t \in \{w_1, ..., w_{C+1}\}^{H\times W\times Z}$ for the current frame. 
Here, $C+1$ includes $C$ semantic classes with an occupancy state in the scene, while $w_{(\cdot)}$ represents the voxel grid.
%

\subsection{Distillation Architecture}
\noindent\textbf{Framework overview.}
The overall architecture is illustrated in Figure~\ref{fig2:framework}, consisting of teacher and student networks. The teacher network takes both LiDAR and multi-view images as input, while the student network solely utilizes multi-view images. Both branches are supervised by ground truth occupancy, and the distillation constraints are applied between 3D occupancy predictions and rendered results.

\noindent\textbf{Camera-based student.} Our student network takes multi-frame multi-view images as input and first extracts the feature using an image backbone. To leverage the benefits of Bird's Eye View (BEV) perception, we apply pixel-wise depth estimation on image features and then project them from the perspective view into a 3D volume via the view-transform operation proposed in \cite{huang2021bevdet}, forming a low-level volume feature.
Moreover, to introduce the temporal information in our model, we adopt the technique proposed in \cite{li2022bevstereo}, which dynamically warps and fuses the historical volume feature and produces a fused feature.
To obtain more fine-grained predicted shapes, the volume feature is fed into an occupancy decoder to generate the prediction. 
%

\noindent\textbf{Multi-modal teacher.}
Inspired by LiDAR-based detectors presented in \cite{shi2020points}, the unstructured point clouds are scattered into pillars~\cite{lang2019pointpillars}.
Subsequently, the volume features are extracted by SECOND and SECOND-FPN~\cite{yan2018second}.
Building upon the success of LiDAR-camera-based BEV detectors, as presented in \cite{liu2023bevfusion}, we further concatenate features from two modalities and process the result with a fully convolutional network to produce the fused features. 
Finally, a similar occupancy decoder is applied to the fused feature, resulting in the prediction of occupancy.

\begin{figure}[t]
	\begin{center}
		\includegraphics[width=\linewidth]{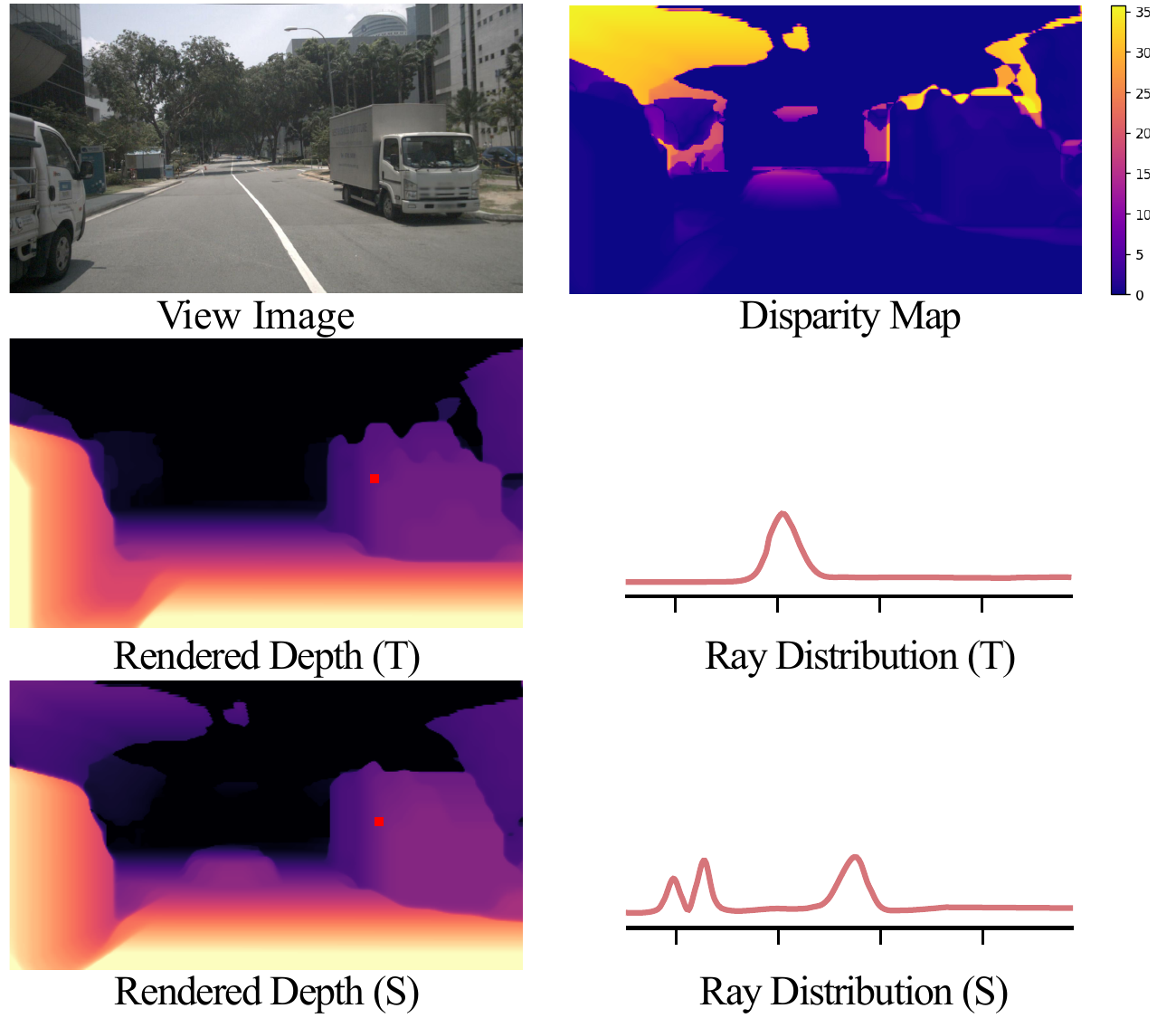}
	\end{center}
	\caption{\textbf{The analysis of rendered depths.} Although the rendered depths of teacher (T) and student (S) are similar, especially for the foreground objects, their ray termination distribution shows a great disparity.}
	\label{fig3:rdc}
\end{figure}

\subsection{Rendering Assisted Distillation}
\noindent\textbf{Volume rendering.}
In this paper, we adopt the volume rendering technique as proposed in NeRF~\cite{mildenhall2021nerf} to obtain depth and semantic maps for knowledge distillation.
By incorporating camera intrinsic and external parameters, we are able to compute the corresponding 3D ray for each pixel in the 2D image. 
%
%
After that, we employ the volume rendering technique to perform a weighted sum on the sampled points along the ray, thereby calculating the predicted depths and semantics in perspective views.
Given $N_p$ sampled points $\{p_i=(x_i, y_i,z_i))\}_{i=1}^{N_p}$ along the ray in pixel $(u, v)$, the rendered depth $\hat{d}$ and semantic logits $\hat{s}$ at this pixel can be calculated via

\begin{align} \label{eq1}
	T_i &= \mathrm{exp}(\sum\nolimits_{j=1}^{i-1}\sigma(p_j) \delta_j), \\
	\hat{d}(u, v) &=  \sum\nolimits_{i=1}^{N_p} T_i (1-\mathrm{exp}(-\sigma(p_i) \delta_i)){d}(p_i),\\
	\hat{s}(u, v) &=  \sum\nolimits_{i=1}^{N_p} T_i (1-\mathrm{exp}(-\sigma(p_i) \delta_i)){s}(p_i),
\end{align}
where $d(\cdot)$, $\sigma(\cdot)$ and $s(\cdot)$ are distance, volume density and semantic of the sampled point, respectively.
Since the occupancy network will predict the occupancy probability and semantics, we can easily obtain $\sigma(p_i)$ and $s(p_i)$ by scattering the voxel predictions into the corresponding sampled point $ p_i$.
Moreover, $\delta_i = {d}(p_{i+1}) - {d}(p_i)$ is the distance between two adjacent sampled points.
Finally, we obtain depth and semantic maps in $i$-th perspective view through collecting results from all pixels, \textit{i.e.,} $\mathcal{S}_i = \{\hat{s}(u, v)~|~u \in [1, H], v\in[1, W] \}$ and $\mathcal{D}_i = \{\hat{d}(u, v)~|~u \in [1, H], v\in[1, W]\}$, where $(H, W)$ is the size of view image. 
To facilitate the definition, we respectively denote rendered depth and semantics results from teacher and student as $\mathcal{D}^{T/S} = \{\mathcal{D}^{T/S}_1,...,\mathcal{D}^{T/S}_n\}$ and $\mathcal{S}^{T/S} = \{\mathcal{S}^{T/S}_1,...,\mathcal{S}^{T/S}_n\}$, where $n$ is the number of views.

\noindent\textbf{Rendered depth consistency.}
After acquiring the rendered depth, a simplistic approach involves directly imposing constraints between the output of teacher and student models. 
However, this approach is a hard constraint, and the differences in rendered depths between the teacher and student models are typically within a narrow range. 
To address this issue, we propose an innovative approach that aligns the ray termination distribution during the volume rendering process.
As shown in Figure~\ref{fig3:rdc}, we plot ray distribution over the distance traveled by the ray.
Although the rendered depths of the two models are quite similar, their ray distribution shows a great discrepancy.
When a ray traverses through single objects (the red point), we find that the ray termination distribution of the teacher model is typically unimodal, while that of the student exists multiple peaks.
Aligning this distribution makes the student model tend to predict a similar latent distribution as the teacher model.
Finally, rendered depth consistency (RDC) loss $\mathcal{L}_{rdc}$ is formulated as 

\begin{align} \label{eq2}
	\mathcal{R}^{(\cdot)}_{(u, v)} &=  \{T_i (1-\mathrm{exp}(-\sigma(p_i) \delta_i))\}_{i=1}^{N_p}, \\
	\mathcal{L}_{rdc} &= \frac{1}{HW}\sum_{u=1}^{H}\sum_{v=1}^{W}\mathcal{D}_{KL}(\mathcal{R}_{(u, v)}^{\mathrm{teacher}}||\mathcal{R}_{(u, v)}^{\mathrm{student}}).
\end{align}
Here, $T_i$ is calculated as Eqn.~\eqref{eq1}.
The notation $\mathcal{R}^{\mathrm{teacher}}$ and $\mathcal{R}^{\mathrm{student}}$ respectively denote the ray distribution of the teacher and student networks, which are aligned through KL divergence $\mathcal{D}_{KL}(\cdot || \cdot)$.


\begin{figure}[t]
	\begin{center}
		\includegraphics[width=0.95\linewidth]{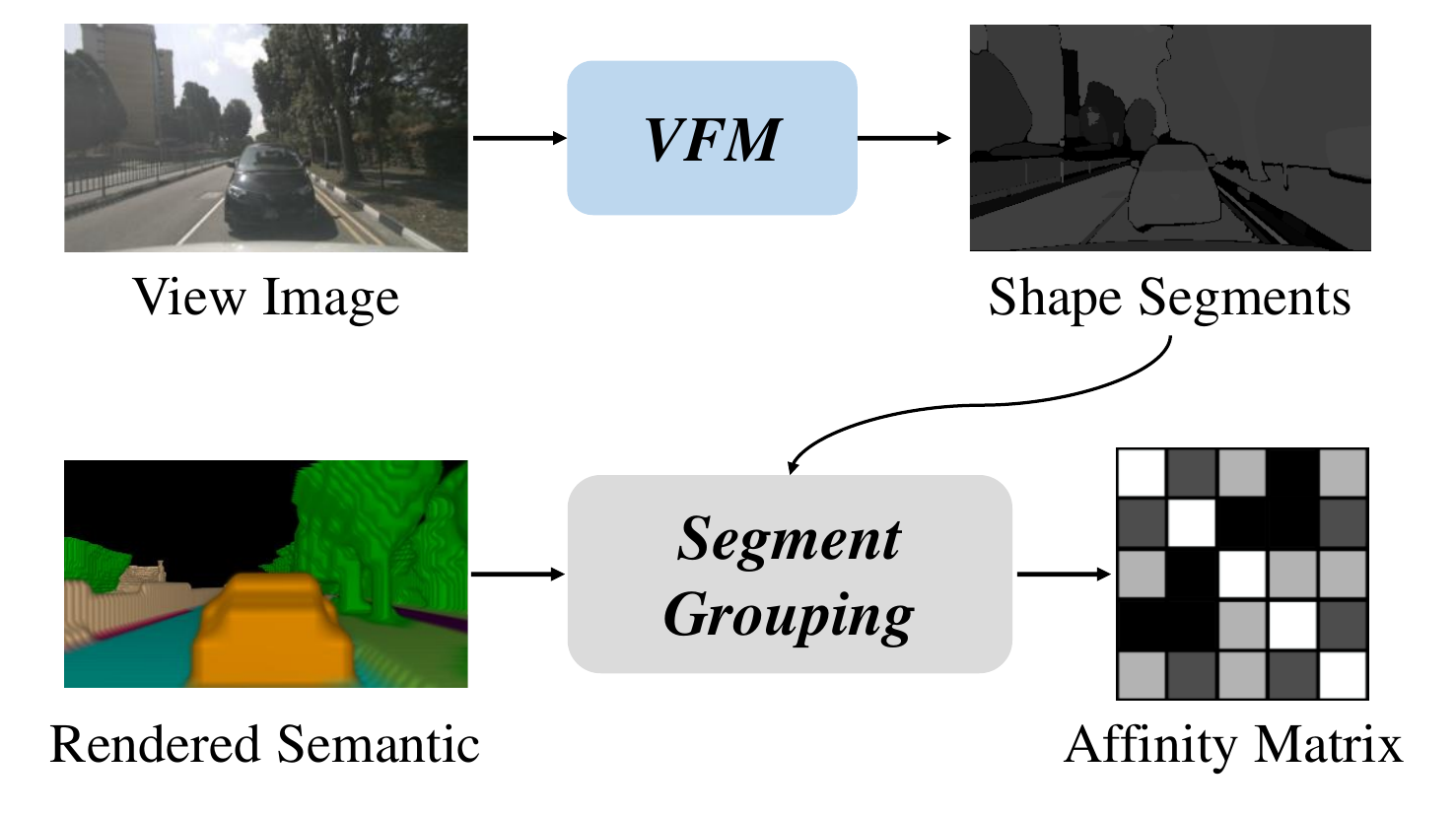}
	\end{center}
	\caption{\textbf{The generation of affinity matrix.} We first adopt visual foundation model (VFM), \textit{i.e.,} SAM, to extract segments into the original image. After that, we conduct segment grouping in rendered semantic features in each segment, obtaining the affinity matrix.}
	\label{fig3b:rsc}
\end{figure}

\begin{table*}[ht]
	\footnotesize
	\setlength{\tabcolsep}{0.0025\linewidth}
	
	\newcommand{\classfreq}[1]{{~\tiny(\nuscenesfreq{#1}\%)}}  %
	\begin{center}
		\resizebox{\textwidth}{!}{
			\begin{tabular}{l|c|c| c c c c c c c c c c c c c c c c c}
				\toprule
				Method
				& \makecell{Image \\ Backbone} & mIoU
				& \rotatebox{90}{\textcolor{nothers}{$\blacksquare$} others}
				
				& \rotatebox{90}{\textcolor{nbarrier}{$\blacksquare$} barrier}
				
				& \rotatebox{90}{\textcolor{nbicycle}{$\blacksquare$} bicycle}
				
				& \rotatebox{90}{\textcolor{nbus}{$\blacksquare$} bus}
				
				& \rotatebox{90}{\textcolor{ncar}{$\blacksquare$} car}
				
				& \rotatebox{90}{\textcolor{nconstruct}{$\blacksquare$} const. veh.}
				
				& \rotatebox{90}{\textcolor{nmotor}{$\blacksquare$} motorcycle}
				
				& \rotatebox{90}{\textcolor{npedestrian}{$\blacksquare$} pedestrian}
				
				& \rotatebox{90}{\textcolor{ntraffic}{$\blacksquare$} traffic cone}
				
				& \rotatebox{90}{\textcolor{ntrailer}{$\blacksquare$} trailer}
				
				& \rotatebox{90}{\textcolor{ntruck}{$\blacksquare$} truck}
				
				& \rotatebox{90}{\textcolor{ndriveable}{$\blacksquare$} drive. suf.}
				
				& \rotatebox{90}{\textcolor{nother}{$\blacksquare$} other flat}
				
				& \rotatebox{90}{\textcolor{nsidewalk}{$\blacksquare$} sidewalk}
				
				& \rotatebox{90}{\textcolor{nterrain}{$\blacksquare$} terrain}
				
				& \rotatebox{90}{\textcolor{nmanmade}{$\blacksquare$} manmade}
				
				& \rotatebox{90}{\textcolor{nvegetation}{$\blacksquare$} vegetation}
				
				\\
				\midrule
				\multicolumn{20}{c}{Performances on Validation Set} \\
				\midrule
				MonoScene  & R101-DCN & 6.06 & 1.75 & 7.23 & 4.26 & 4.93 & 9.38 & 5.67 & 3.98 & 3.01 & 5.90 & 4.45 & 7.17 & 14.91 & 6.32 & 7.92 & 7.43 & 1.01 & 7.65\\
				CTF-Occ  & R101-DCN & 28.53 & 8.09 & 39.33 & 20.56 & 38.29 & 42.24 & 16.93 & 24.52 & 22.72 & 21.05 & 22.98 & 31.11 & 53.33 & 33.84 & 37.98 & 33.23 & 20.79 & 18.00\\
				BEVFormer  & R101-DCN & 39.24 & 10.13 & 47.91 & 24.90 & 47.57 & 54.52 & 20.23 & 28.85 & 28.02 & 25.73 & 33.03 & 38.56 & 81.98 & 40.65 & 50.93 & 53.02 & 43.86& 37.15  \\
				PanoOcc  & R101-DCN & {42.13} & 11.67 & 50.48 & 29.64 & 49.44 & 55.52 & 23.29 & \textbf{33.26} & 30.55 & 30.99 & 34.43 & 42.57 & \textbf{83.31} & 44.23 & 54.40 & 56.04 & 45.94 & 40.40  \\
				BEVDet$\dagger$  & Swin-B & 42.02 & 12.15 & 49.63 & 25.10 & 52.02 & 54.46 & 27.87 & 27.99 & 28.94 & 27.23 & 36.43 & 42.22 & 82.31 & 43.29 & 54.62 & 57.90 & 48.61 & 43.55  \\ 
				
				\midrule
                Baseline (ours) & Swin-B & 44.14 & \textbf{13.39}& 52.20 & \textbf{31.43} & 52.01 & 56.70 & \textbf{30.66} & 32.95 &31.56 &\textbf{31.31} &39.87 &44.64 &82.98 &\textbf{44.97} &\textbf{55.43} & \textbf{58.90} &48.43&42.99\\
				RadOcc  (ours)  & Swin-B & \textbf{46.06} &9.78 &\textbf{54.93} &20.44 &\textbf{55.24} &\textbf{59.62} &30.48 &28.94 &\textbf{44.66} &28.04 &\textbf{45.69} &\textbf{48.05} &81.41 &39.80 &52.78 &56.16 &\textbf{64.45} &\textbf{62.64}\\
				{\color{gray} Teacher (ours)}    & {\color{gray} Swin-B} & {\color{gray} 49.38} & {\color{gray} 10.93}& {\color{gray} 58.23} & {\color{gray} 25.01} & {\color{gray} 57.89} & {\color{gray} 62.85} & {\color{gray} 34.04} & {\color{gray} 33.45} & {\color{gray} 50.07} & {\color{gray} 32.05} & {\color{gray} 48.87} & {\color{gray} 52.11} & {\color{gray} 82.9} & {\color{gray} 42.73} & {\color{gray} 55.27} & {\color{gray} 58.34} & {\color{gray} 68.64} & {\color{gray} 66.01} \\
				\midrule

				\multicolumn{20}{c}{Performances on  3D Occupancy Prediction Challenge} \\
				\midrule
				BEVFormer & R101-DCN  & 23.70 & 10.24&36.77&11.70&29.87&38.92&10.29&22.05&16.21&14.69&27.44&33.13&48.19&33.10&29.80&17.64&19.01&13.75 \\
				SurroundOcc$\dagger$ & R101-DCN & 42.26 &11.7&50.55&32.09&41.59&57.38&27.93&38.08 &30.56&29.32&48.29&38.72&80.21&48.56&53.20&47.56&46.55&36.14 \\
				BEVDet$\dagger$ & Swin-B &42.83&18.66&49.82&31.79&41.90&56.52&26.74&37.31&30.01&31.33&48.18&38.59&80.95&50.59&53.87&49.67&46.62&35.62 \\
				PanoOcc-T$\star$ & Intern-XL & 47.16 & \textbf{23.37} & 50.28 & 36.02 & 47.32 & 59.61 & 31.58 & 39.59 & 34.58 & 33.83 & 52.25 & 43.29 & \textbf{83.82} & \textbf{55.81} & \textbf{59.41} & 53.81 & 53.48 & 43.61 \\
				\midrule
				Baseline-T  (ours) & Swin-B & 47.74 &22.88&50.74&\textbf{41.02}&\textbf{49.39}&55.40&33.41&45.71&38.57&35.79&48.94&44.40&{83.19}&{52.26}&59.09&\textbf{55.83}&51.35&43.54\\
				RadOcc-T (ours)  & Swin-B & \textbf{49.98}&21.13&\textbf{55.17}&39.31&48.99&\textbf{59.92}&\textbf{33.99}&\textbf{46.31}&\textbf{43.26}&\textbf{39.29}&\textbf{52.88}&\textbf{44.85}&83.72&53.93&{59.17}&{55.62}&\textbf{60.53}&\textbf{51.55}\\
				
				{{\color{gray} Teacher-T (ours)}}  & {{\color{gray} Swin-B}} & {\color{gray} 55.09} & {\color{gray} 25.94}& {\color{gray} 59.04} & {\color{gray} 44.93} & {\color{gray} 57.95} & {\color{gray} 63.70} & {\color{gray} 38.89} & {\color{gray} 52.03} & {\color{gray} 53.21} & {\color{gray} 42.16} & {\color{gray} 59.90} & {\color{gray} 50.45} & {\color{gray} 84.79} & {\color{gray} 55.70} & {\color{gray} 60.83} & {\color{gray} 58.02} & {\color{gray} 67.66} & {\color{gray} 61.40} \\
				\bottomrule
			\end{tabular}
		}
	\end{center}
    \caption{\textbf{3D occupancy prediction performance on the Occ3D.} $\dagger$ denotes the performance reproduced by official codes. $\star$~means the results provided by authors. `-T' represents results through test-time augmentation (TTA). Please note that our visual model achieves a benchmark ranking of \textbf{Top-4} on 16/08/2023, outperforming all previously published methods. }
	\label{tab:camera_occ}
\end{table*}

\noindent\textbf{Rendered semantic consistency.}
%
%
Besides simply using KL divergence to align the semantic logits, we also leverage the strengths of vision foundation models (VFMs)~\cite{kirillov2023segment} to perform a segment-guided affinity distillation (SAD). 
Specifically, we first employ the VFM to over-segment patches using the original view images as input, as illustrated in Figure~\ref{fig3b:rsc}.
With the rendered semantic features from both the teacher and student networks, \textit{i.e.,} $\mathcal{S}^{T}$, $\mathcal{S}^{S} \in \mathbb{R}^{H \times W \times C}$, we can divide the rendered semantics into several groups based on the indices of aforementioned patches.
After that, an average pooling function is applied within each group, extracting multiple teacher and student semantic embedding, \textit{i.e.,} $\mathcal{E}^T \in \mathbb{R}^{M \times C} $ and $\mathcal{E}^S \in \mathbb{R}^{M \times C} $.
Here, $M$ is the number of patches generated by the VFM.
Inspired but different from the previous work~\cite{hou2022point}, we calculate an affinity matrix $\mathcal{C}^{(\cdot)}$  according to the above segments for the further distillation:

\begin{align}
	\mathcal{C}_{i,j,r} = \frac{\mathcal{E}(i,r), \mathcal{E}(j,r)}{||\mathcal{E}(i)||_2 ||\mathcal{E}(j)||_2}.
	\label{eq:loss_slidr}
\end{align}
The affinity score captures the similarity of each segment  of semantic embedding and it can be taken as the high-level structural knowledge to be learned by the student. After that, the final RSC loss is a linear combination of affinity distillation loss and KL divergence between rendered semantics:

\begin{align}
	\mathcal{L}_{sad} &=  \sum_{r=1}^C \sum_{i=1}^{M} \sum_{j=1}^M ||C^T_{i,j,r} - C^S_{i,j,r}||^2_2, \\
	\mathcal{L}_{rsc} &=  \mathcal{L}_{sad}/CM^2 + \omega \mathcal{D}_{KL}(\mathcal{S}^{{T}}||\mathcal{S}^{{S}}),
	\label{eq:loss_rsc}
\end{align}
where $C^T$ and $C^S$ are affinity matrices of teacher and student networks, and $\omega$ is a hyperparameter in our experiment.

\begin{table*}[ht]
	\footnotesize
	\setlength{\tabcolsep}{0.0045\linewidth}
	
	\newcommand{\classfreq}[1]{{~\tiny(\nuscenesfreq{#1}\%)}}  %
	\begin{center}
		
		\begin{tabular}{l|c|c|c | c c c c c c c c c c c c c c c c}
			\toprule
			Method
			& \makecell{Input \\ Modality}& \makecell{Image \\ Backbone} & mIoU
			& \rotatebox{90}{\textcolor{nbarrier}{$\blacksquare$} barrier}
			
			& \rotatebox{90}{\textcolor{nbicycle}{$\blacksquare$} bicycle}
			
			& \rotatebox{90}{\textcolor{nbus}{$\blacksquare$} bus}
			
			& \rotatebox{90}{\textcolor{ncar}{$\blacksquare$} car}
			
			& \rotatebox{90}{\textcolor{nconstruct}{$\blacksquare$} const. veh.}
			
			& \rotatebox{90}{\textcolor{nmotor}{$\blacksquare$} motorcycle}
			
			& \rotatebox{90}{\textcolor{npedestrian}{$\blacksquare$} pedestrian}
			
			& \rotatebox{90}{\textcolor{ntraffic}{$\blacksquare$} traffic cone}
			
			& \rotatebox{90}{\textcolor{ntrailer}{$\blacksquare$} trailer}
			
			& \rotatebox{90}{\textcolor{ntruck}{$\blacksquare$} truck}
			
			& \rotatebox{90}{\textcolor{ndriveable}{$\blacksquare$} drive. suf.}
			
			& \rotatebox{90}{\textcolor{nother}{$\blacksquare$} other flat}
			
			& \rotatebox{90}{\textcolor{nsidewalk}{$\blacksquare$} sidewalk}
			
			& \rotatebox{90}{\textcolor{nterrain}{$\blacksquare$} terrain}
			
			& \rotatebox{90}{\textcolor{nmanmade}{$\blacksquare$} manmade}
			
			& \rotatebox{90}{\textcolor{nvegetation}{$\blacksquare$} vegetation}
			
			\\
			\midrule

			PolarNet & LiDAR &- & 69.4 &72.2&16.8&77.0&86.5&51.1&69.7&64.8&54.1&69.7&63.5&96.6&67.1&77.7&72.1&87.1&84.5\\
			Cylinder3D & LiDAR &- & 77.2&82.8&29.8&84.3&89.4&63.0&79.3&77.2&73.4&84.6&69.1&97.7&70.2&80.3&75.5&90.4&87.6\\
			2DPASS & LiDAR & -  & 80.8 &81.7&55.3&92.0&91.8&73.3&86.5&78.5&72.5&84.7&75.5&97.6&69.1&79.9&75.5&90.2&88.0\\
			
			\midrule
			TPVFormer  & Camera & R50-DCN & 59.2 &65.6&15.7&75.1&80.0&48.8&43.1&44.3&26.8&72.8&55.9&92.3&53.7&61.0&59.2&79.7&75.6 \\
   			BEVDet$\dagger$ & Camera & Swin-B & 65.2 &  31.3&\textbf{63.9}&74.6&79.1&51.5&59.8&63.4&56.2&74.7&59.8&92.8&61.4&69.5&65.7&84.1&82.9 \\
			TPVFormer (BL) & Camera & R101-DCN & 69.4 &\textbf{74.0}&27.5&\textbf{86.3}&85.5&\textbf{60.7}&68.0&62.1&49.1&\textbf{81.9}&68.4&94.1&59.5&66.5&63.5&83.8&79.9\\ %

			\midrule
			RadOcc (ours)& Camera & R101-DCN & \textbf{71.8} & 49.1 & 34.2 & 84.5 & \textbf{85.8} & 59.2 & \textbf{70.3} & \textbf{71.4} & \textbf{62.5} & 79.7 & \textbf{69.0} & \textbf{95.4} & \textbf{66.2} & \textbf{75.1} & \textbf{72.0} & \textbf{87.4} & \textbf{86.0}\\
			{{\color{gray} Teacher (ours)}}  & {{\color{gray} Cam+Li}}  &{{\color{gray} R101-DCN}} &{\color{gray} 75.2} &{\color{gray} 62.7} &{\color{gray} 33.2} &{\color{gray} 88.7}  &{\color{gray} 88.8} &{\color{gray} 64.6} &{\color{gray} 78.1} &{\color{gray} 74.1}  &{\color{gray} 65.0} &{\color{gray} 83.1} &{\color{gray} 72.2} &{\color{gray} 96.5} &{\color{gray} 68.3}  &{\color{gray} 77.6} &{\color{gray} 74.4}  &{\color{gray} 88.7} &{\color{gray} 87.1}\\
			
			\bottomrule
		\end{tabular}
	\end{center}
    \caption{\textbf{LiDAR semantic segmentation results on nuScenes test benchmark.} $\dagger$ denotes the performance is reproduced by official codes. Our method achieves state-of-the-art performance in camera-based methods. BL denotes the baseline method.
	}
	\label{tab:lidar_seg}
\end{table*}

\section{Experiments}
\subsection{Dataset ane Metric}
\noindent \noindent \textbf{Dataset.}  We evaluate our proposed method on {nuScenes}~\cite{caesar2020nuscenes} for sparse prediction and
{Occ3D}~\cite{tian2023occ3d} for dense prediction.
The data descriptions are provided in supplementary material.

%
%


\begin{figure*}[t]
	\begin{center}
		\includegraphics[width=1.0\linewidth]{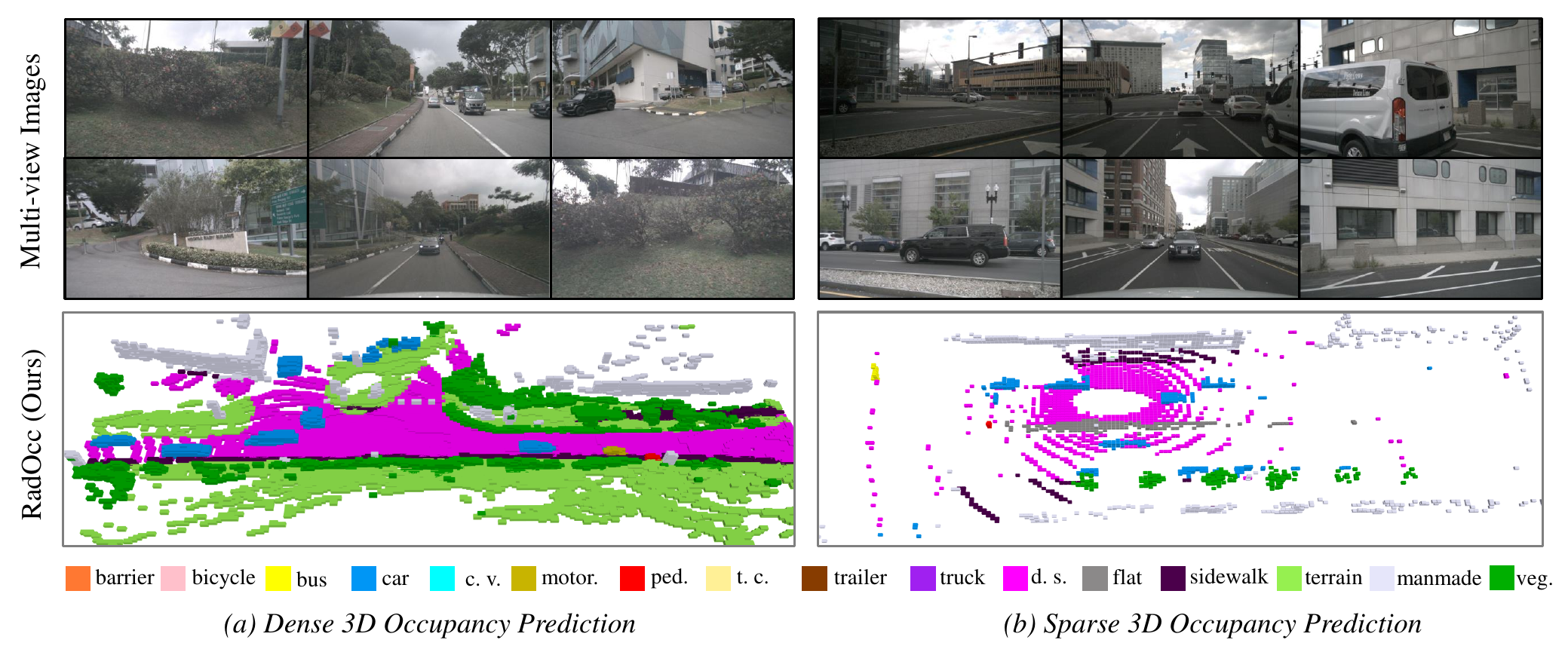}
	\end{center}
	\caption {\textbf{Qualitative results on Occ3D and nuScenes validation sets.} RadOcc takes multi-view images as input and produces voxel predictions. More visualization comparisons can be found in the supplementary materials.}
	\label{fig4:visualization}
\end{figure*}

\noindent \textbf{Evaluation metrics.} 
Our study presents an independent evaluation of the model's performance in both dense and sparse prediction tasks. 
Specifically, for {dense prediction}, we conduct experiments on the Occ3D dataset, which quantifies the mean Intersection over Union (mIoU) for 17 semantic categories within the camera's visible region. 
On the other hand, for {sparse prediction}, we train the model with single-sweep LiDAR and assess the model's performance on the nuScenes-lidarseg benchmark, which measures the mIoU for 16 semantic categories, with the `others' category being treated as `ignored'.

\subsection{Experimental Settings}
\noindent \textbf{Implementation.}
For the dense prediction, we follow the setting of BEVDet~\cite{huang2021bevdet} and use Swin Transformer~\cite{liu2021swin} as the image backbone. 
We adopt the semantic scene completion module proposed in \cite{yan2021sparse} as our occupancy decoder, which contains several 3D convolutional blocks to learn a local geometry representation. Afterward, the features from different blocks are concatenated to aggregate information. Finally, a linear projection is utilized to map the feature into $C+1$ dimensions.
Since the challenging nature of the Occ3D test benchmark, we utilize {8 historical frames} for temporal encoding and use 3 frames on the validation set.
For the sparse prediction, we use previous art TPVFormer~\cite{huang2023tri} as our baseline.
The rendered size of the network is configured to $384\times 704$. To speed up the rendering and reduce memory usage, we randomly sample 80,000 rays during each step.

\subsection{Results and Analysis}

\noindent\textbf{Dense Prediction.} To evaluate the performance of dense 3D occupancy prediction, we compare our proposed method with current state-of-the-art approaches on the Occ3D dataset~\cite{tian2023occ3d}, including the validation set and online benchmark.
The upper part of Table~\ref{tab:camera_occ} presents the validation set results, where all methods are trained for 24 epochs. 
Specifically, we compare our approach with  MonoScene~\cite{cao2022monoscene},  BEVFormer~\cite{li2022bevformer},  CTF-Occ~\cite{tian2023occ3d} and PanoOcc~\cite{wang2023panoocc}, which all employ the ResNet101-DCN~\cite{dai2017deformable} initialized from FCOS3D~\cite{wang2021fcos3d} checkpoint as the image backbone. 
Additionally, we report the results of BEVDet~\cite{huang2021bevdet} that uses the same image backbone as ours.
Our baseline model, trained from scratch, already outperforms prior state-of-the-art methods. However, by leveraging our proposed distillation strategy, we achieve significantly better occupancy results in terms of mIoU.

The lower part of Table~\ref{tab:camera_occ} presents the results on the 3D occupancy prediction challenge, where our proposed method achieves state-of-the-art performance and outperforms all previously published approaches by a large margin.
Note that though PanoOcc~\cite{wang2023panoocc} adopts a stronger image backbone, \textit{i.e.,} InternImage-XL~\cite{wang2022internimage}, the results of them are still lower than ours, especially for the foreground objects with challenge nature.
The visualization results for both dense and sparse prediction are shown in Figure~\ref{fig4:visualization}. More visualization results can be found in the supplementary material.

\begin{table}[t]
	\footnotesize
	\centering
	\begin{tabular}{l|c|cc}
		\toprule
		Method &  Consistency & mIoU & Gains\\\hline
		BEVDet (baseline) & - & 36.10 &-\\
		Hinton \textit{et al.}& Prob. & 37.00 & { +0.90} \\
		Hinton  \textit{et al.} &  Feature & 35.89  & { -0.21} \\
		BEVDistill &  Prob. + Feature &35.95 & { -0.15}\\ \midrule
		RadOcc (ours) & Render & 37.98 & {+1.88}\\
		RadOcc (ours) & Prob. + Render & \textbf{38.53} & {+2.43}\\ 
		
		\bottomrule
	\end{tabular}
     \caption{\textbf{Comparison for knowledge distillation.} The results are obtained on Occ3D. To speed up the evaluation, we take BEVDet~\cite{huang2021bevdet} with ResNet50 image backbone as our baseline. $\dagger$: Since there is no object-level prediction, we replace the sparse distillation of BEVDistill~\cite{chen2022bevdistill} with logits distillation.}
	\label{tab:kt}
\end{table}%

\noindent\textbf{Sparse Prediction.} To evaluate the effectiveness of model using sparse LiDAR supervision, we evaluate the performance of our proposed RadOcc model on the nuScenes LiDAR semantic segmentation benchmark. 
Our results, as shown in Table~\ref{tab:lidar_seg}, demonstrate a significant improvement over the baseline TPVFormer~\cite{huang2023tri} and outperform previous camera-based occupancy networks such as BEVDet~\cite{huang2021bevdet}.
Surprisingly, our method even achieves comparable performance with some LiDAR-based semantic segmentation methods~\cite{zhang2020polarnet,zhou2020cylinder3d}. 
It should be noted that since we use voxelized single-sweep LiDAR as supervision, where the geometric details in data may be lost during the voxelization, the results of a multi-modal teacher network may not achieve comparable performance with state-of-the-art LiDAR-based methods~\cite{yan20222dpass}.

\noindent\textbf{Comparison for knowledge distillation.}
To further validate the efficacy of our proposed methodology upon previous teacher-student architectures, we conduct a comparative analysis of RadOcc with conventional knowledge transfer techniques as presented in Table~\ref{tab:kt}. 
To facilitate the experimentation process, we choose BEVDet~\cite{huang2021bevdet} with ResNet50 image backbone as our baseline, and all methods are trained with the same strategies for a fair comparison. 
The results in the table indicate that direct application of feature and logits alignment~\cite{hinton2015distilling,chen2022bevdistill} fails to achieve a significant boost on the baseline model, particularly for the former, which results in negative transfer. 
Notably, leveraging rendering-assisted distillation leads to a substantial improvement of 2\% on mIoU. Furthermore, even when applying logit distillation, the model can still enhance the mIoU by 0.6\%.
%

\begin{table}[t]
	\footnotesize
	\centering

	\begin{tabular}{l|cc|cc|c}
		\toprule
		Method & RDC(-) & RDC& SAD &RSC & mIoU\\\hline
		BEVDet  &  & &&& 36.10 \\ 
		\midrule
		Model A & {\color{black} \checkmark} && && { 35.08}\\
		Model B &  &{\color{black} \checkmark} &&& 36.76\\
		Model C &  & &{\color{black} \checkmark}& &37.13\\
		Model D &  & &&{\color{black} \checkmark}& 37.42\\
		
		\midrule
		RadOcc (ours) && {\color{black} \checkmark} &&{\color{black} \checkmark}&  \textbf{37.98}\\

		\bottomrule
	\end{tabular}
 \caption{\textbf{Ablation study on Occ3D.} We use BEVDet with ResNet50 image backbone as our baseline. Here, RDC and RSC are rendered depth and semantic consistency losses. RDC (-) denotes directly aligning the rendered depth map with Scale-Invariant Logarithmic loss.
	}
	\label{tab:abl}%
\end{table}%

\noindent\textbf{Ablation study.}
We conduct an ablation study of rendering distillation in Table~\ref{tab:abl}. Here, BEVDet with ResNet50 image backbone is selected as our baseline model.
\textit{Model A} directly conducts alignment through Scale-Invariant Logarithmic~\cite{eigen2014depth} on rendered depth maps but fails to improve the performance. In contrast, \textit{Model B} aligns the latent distribution of depth rendering and achieves an improvement of 0.7\% in mIoU. 
On the other hand, \textit{Model C} demonstrates the results sorely using segment-guided affinity distillation (SAD) on rendered semantics, which increases the mIoU by 1.0\%.
Applying additional KL divergence between two rendered semantics can boost the performance to 37.42\%.
Finally, when we combine RDC and RSC losses, the model achieves the best result.

In Table~\ref{tab:abl2}, we analyze the design of SAD by replacing its segment with other implementations in \textit{Model~E}. Specifically, when we use super-pixel~\cite{achanta2012slic}, the performance will decrease by about 0.37\%.

\begin{table}[t]
	\footnotesize
	\centering
	\begin{tabular}{l|c|cc}
		\toprule
		Method & Segment & mIoU & Gains\\\hline
		\midrule
		BEVDet w/ RSC &  SAM &  \textbf{37.42} & -\\
		
		\midrule
		Model E & Super Pixel &  37.05 & { -0.37} \\
		
		\bottomrule
	\end{tabular}
    \caption{\textbf{Design analysis of SAD.} We replace the segment extraction strategy with other designs. 
	}
	\label{tab:abl2}
\end{table}%

\section{Conclusion}
In this paper, we present RadOcc, a novel cross-modal knowledge distillation paradigm for 3D occupancy prediction. 
It leverages a multi-modal teacher model to provide geometric and semantic guidance to a visual student model via differentiable volume rendering. 
Moreover, we propose two new consistency criteria, depth consistency loss and semantic consistency loss, to align the ray distribution and affinity matrices between the teacher and student models. 
Extensive experiments on the Occ3D and nuScenes datasets show RadOcc can significantly improve the performance of various 3D occupancy prediction methods. 
Our method achieves state-of-the-art results on the Occ3D challenge benchmark and outperforms existing published methods by a large margin. 
We believe that our work opens up new possibilities for cross-modal learning in scene understanding.

\section{Acknowledgments}
This work was supported by NSFC with Grant No. 62293482, by the Basic Research Project No. HZQB-KCZYZ-2021067 of Hetao Shenzhen HK S$\&$T Cooperation Zone, by Shenzhen General Program No. JCYJ20220530143600001, by Shenzhen-Hong Kong Joint Funding No. SGDX20211123112401002, by Shenzhen Outstanding Talents Training Fund, by Guangdong Research Project No. 2017ZT07X152 and No. 2019CX01X104, by the Guangdong Provincial Key Laboratory of Future Networks of Intelligence (Grant No. 2022B1212010001), by the Guangdong Provincial Key Laboratory of Big Data Computing, The Chinese University of Hong Kong, Shenzhen, by the NSFC 61931024$\&$81922046$\&$61902335, by the Shenzhen Key Laboratory of Big Data and Artificial Intelligence (Grant No. ZDSYS201707251409055), and the Key Area R$\&$D Program of Guangdong Province with grant No. 2018B03033800, by Tencent$\&$Huawei Open Fund.

\bibliography{aaai24}

\let\titleold\title
\renewcommand{\title}[1]{\titleold{#1}\newcommand{\thetitle}{#1}}
\def\maketitlesupplementary
   {
   \newpage
       \twocolumn[
        \centering
        \LARGE
        \textbf{Supplementary Material}
        \vspace{3.0em}
       ] 
   }

\clearpage
\setcounter{section}{1}
\maketitlesupplementary

\begin{figure*}[t]
	\begin{center}
		\includegraphics[width=0.9\linewidth]{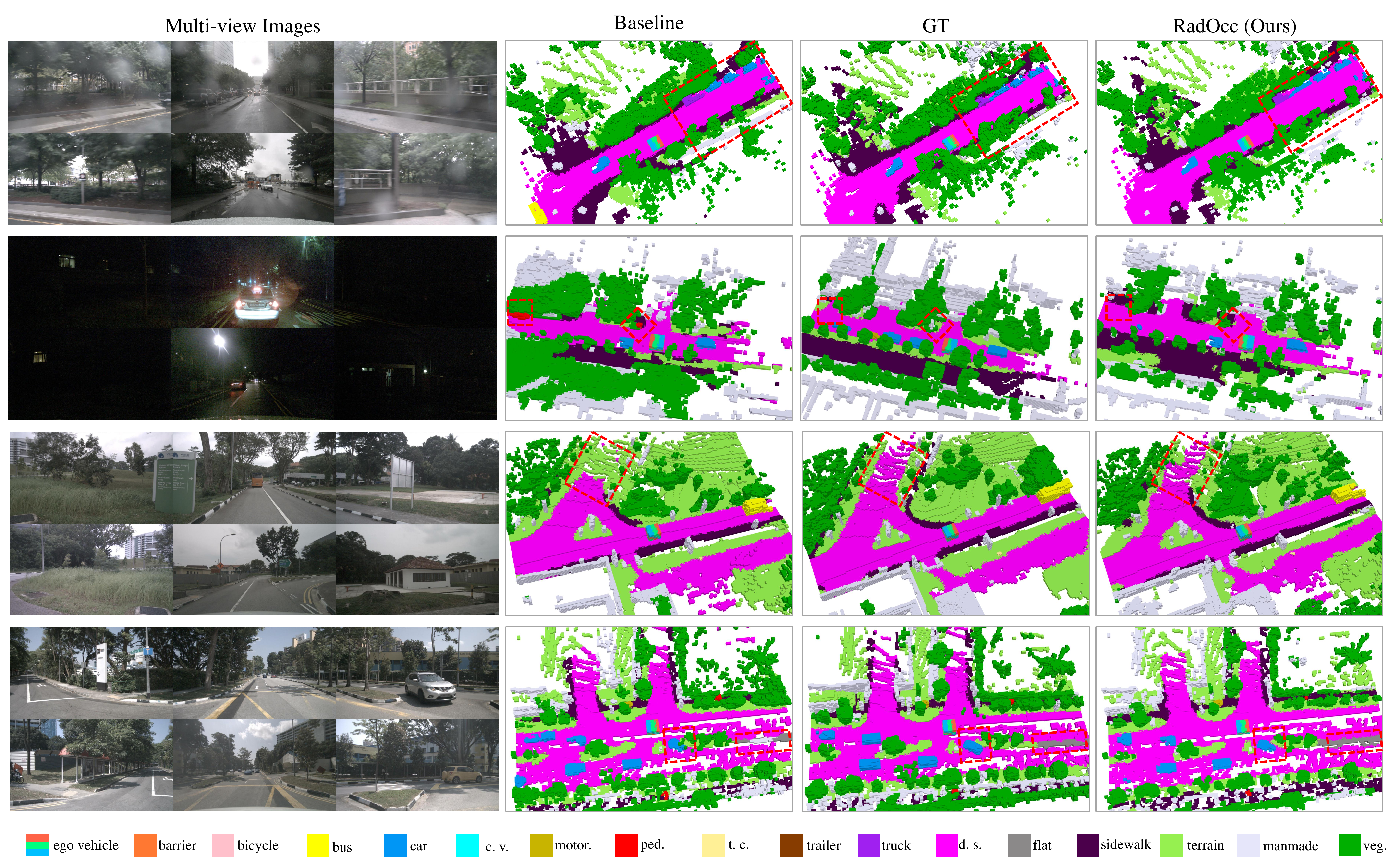}
	\end{center}
	\caption {\textbf{More dense 3D occupancy prediction visualization results on Occ3D validation sets.} We highlight the predicted results by using the red dashed rectangles. Please zoom in for details.}
	\label{fig:comparision}
\end{figure*}

\begin{figure*}[t]
	\begin{center}
		\includegraphics[width=0.9\linewidth]{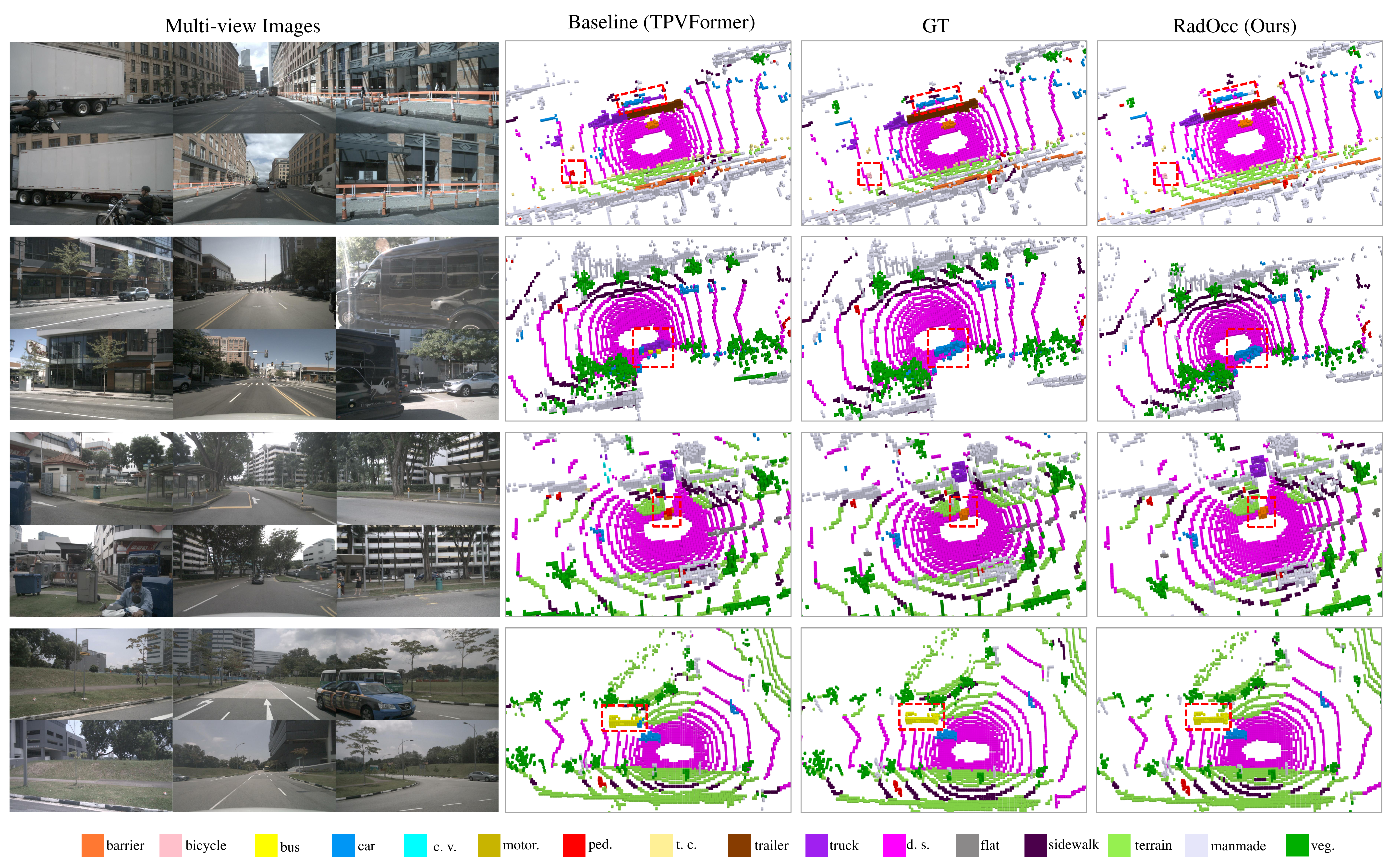}
	\end{center}
	\caption {\textbf{More sparse 3D occupancy prediction visualization results on nuScenes validation sets.} We highlight the predicted results by using the red dashed rectangles. Please zoom in for details.}
	\label{fig:comparision_lidarseg}
\end{figure*}

\section{Overview}
In this supplementary material, we first propose more implementation details about our proposed RadOcc, including dataset descriptions, volume rendering, and training/testing details. 
After that, we provide more ablation studies on the Occ3D dataset, including the model complexity analysis, design analysis of teacher models and the different weight combinations of distillation losses. 
Finally, we demonstrate more visualization results in the last section.

\section{Implementation Details}
\subsection{Dataset Descriptions}
\noindent \textbf{nuScenes}~\cite{caesar2020nuscenes} is composed of a total of 1,000 scenes, which are partitioned into 700 scenes for training, 150 scenes for validation, and 150 scenes for testing.  
The dataset includes RGB images from 6 cameras with a horizontal field of view of 360$^{\circ}$ and point cloud data from a 32-beam LiDAR sensor. 
For the task of semantic segmentation (\textit{i.e.,} nuScenes-lidarseg), key samples are annotated at a frequency of 2Hz with ground truth labels, and every point is annotated with 6 background and 10 foreground classes.

\noindent \textbf{Occ3D}~\cite{tian2023occ3d} is recently proposed for the task of 3D-OP and shares the same train/val/test split with nuScenes. 
Its test split has been structured as part of the CVPR2023 3D Occupancy Prediction Challenge\footnote{\url{https://opendrivelab.com/AD23Challenge.html}}.
The occupancy scope is defined as ranging from $-40m$ to $40m$ for X and Y-axis, and from $-1m$ to $5.4m$ for the Z-axis in the ego coordinate.
The voxel size for the occupancy label is $0.4m$. 
The semantic labels contain 18 categories, including extra `others' and `free' categories compared with nuScenes. 
Moreover, visibility masks are provided for both LiDAR and camera modalities for evaluation.

\subsection{Volume Rendering}
We propose the pseudocode of volume rendering in Alg.~\ref{alg1}. Here, the inputs $V^D \in \mathbb{R}^{X\times Y\times Z}$, $V^S \in \mathbb{R}^{X\times Y\times Z\times C}$, $K=\{k_i\}_{i=1}^N\in \mathbb{R}^{N\times 4\times 4}$ and $T=\{t_i\}_{i=1}^N\in \mathbb{R}^{N\times 4\times 4}$ are density volume, semantic volume, camera intrinsic and extrinsic, respectively. $N$ is the number of perspective cameras. The output of the volume rendering is rendered depth $D \in \mathbb{R}^{N\times H\times W}$ and semantics $S\in \mathbb{R}^{N\times H\times W\times C}$.
It first calculates the origin ($\text{ray}_o$) and direction ($\text{ray}_d$) of each pixel in \textbf{get\_rays} process, where $\text{ray}_o\in \mathbb{R}^{N\times H\times W\times 3}$ and $\text{ray}_d\in \mathbb{R}^{N\times H\times W\times 3}$.    
After that, it generates the sampled points on each ray through \textbf{get\_points} process, in which it uniformly samples a 3D point on each ray after every step\_size.
The distance between each sampled point and pixel origin can be calculated through Euclidean metrics (\textbf{get\_distance}).
Then, we scatter the volume occupancy state and semantic results into each sampled point through \textbf{grid\_sample}.
Finally, we can obtain the rendered depths and semantics through Eqn.~(1)-(3) in the manuscript.

\begin{algorithm}[t]
	\caption{The pseudocode of volume rendering.}              
	\label{alg1}                        
	\begin{algorithmic}
		\renewcommand{\algorithmicrequire}{\textbf{Input:}}
		\renewcommand{\algorithmicensure}{\textbf{Output:}}
		
		\REQUIRE $V^D$, $V^S$, $K$, $T$, $H$, $W$, step\_size        
		\ENSURE$D$, $S$  
		\STATE \textit{\# get ray origin and direction of each pixel}
		\STATE$\text{rays}_o, \text{rays}_d \leftarrow \textbf{get\_rays}(K,~T,~H,~W)$   
		\STATE \textit{\# get sampled points on each ray}
		\STATE  $\mathcal{P} \leftarrow \textbf{get\_points}(\text{rays}_o,~\text{rays}_d,~$step\_size$)$  
		\STATE \textit{\# get the distance between the sampled point and ray origin}
		\STATE $\text{dist} \leftarrow \textbf{get\_distance}(\text{rays}_o,~\text{rays}_p)$   
		\STATE \textit{\# inject density and semantic on each sample point}
		\STATE $\mathcal{P}^D \leftarrow \textbf{grid\_sample}(V^D,~\mathcal{P})$
		\STATE $\mathcal{P}^S \leftarrow \textbf{grid\_sample}(V^S,~\mathcal{P})$
		\STATE \textit{\# calculate interval of each sampled point pair}
		\STATE $\text{delta} \leftarrow \text{dist}[...,1:] - \text{dist}[...,:-1]$
		\STATE \textit{\# Eqn.~(1)-(3) in manuscript}
		\STATE $D, S \leftarrow \textbf{Render}(\mathcal{P}^D,~\mathcal{P}^S,~\text{delta},~\text{dist}$)

		\RETURN $D$, $S$
	\end{algorithmic}
\end{algorithm}

\subsection{Training and Inference}
\textbf{Training.} We train all models with AdamW optimizer, in which gradient clip is utilized with a learning rate of $1\times10^{-4}$ and a weight decay of $1\times10^{-2}$, a total batch size of 32 on multiple GPUs.
For the ResNet-based image-view encoder, for example, the experiments for all ablation studies, the input image shape is 384$\times$704, and a step learning rate policy is leveraged with the linear warm-up strategy in the first 200 iterations and a warm-up ratio of 0.001.
With respect to Swin Transformer-based image-view encoder, the input image shape is 512$\times$1408, and we leverage a cyclic policy, which linearly increases the learning rate from 1e-4 to 1e-3 in the first 40\% schedule and linearly decreases the learning rate from 1e-3 to 0 in the remaining epochs.
We train all models with a maximum of 24 epochs by default.

During training, we first train the teacher model and freeze it when training the student model with the distillation losses.

\noindent\textbf{Data Augmentation.} We apply the data augmentation on both the image space and BEV space following the settings of BEVDet~\cite{huang2021bevdet}. For the image view space, we apply the random scaling with a range of [0.94, 1.11], random rotating with a range of [$-5.4^{\circ}$, $5.4^{\circ}$], and random flipping with a probability of 0.5. After that, the images are padded and cropped to the needed input size.
In the BEV space, we augment the BEV feature and the target 3D semantic occupancy voxels by random flipping of X and Y axes with a probability of 0.5.

\noindent\textbf{Test-Time Augmentation.} In order to further improve the performance, we apply a combination of Test-Time Augmentations (TTAs) before submitting the results to the Occ-3D challenge benchmark.
Specifically, we use the flipping performed horizontally or vertically at both the image and BEV level, and the extra scaling augmentations at the image input with 0.96 and 1.04.
The final prediction is obtained by computing the mean of all the results.
The TTAs in our experiments could have about +1 gain. 
%

\section{More Experiment Results}



\noindent\textbf{Model Complexity.}
The speed and parameter analysis in Table~\ref{supp:tab1} demonstrates the benefits of cross-modal distillation. 
We test different methods in the same server.
Specifically, we set BEVDet for dense occupancy prediction as our baseline (BL), whose training and inference times per sample are 1.74 and 1.00 seconds, respectively.
Although extra LiDAR input improves performance, it increases the training and inference times by 2.03 and 2.33 seconds, respectively, and introduces extra 9.48M parameters.
In contrast, leveraging our rendering-assisted distillation effectively boosts performance by 2\% mIoU while keeping the inference speed and complexity unchanged.

\begin{table}[t]
	\centering
        
	\small{
		\begin{tabular}{l|ccc|c} 
			\hline
			Model & Training & Inference & \#Param &mIoU \\
			\hline
			BEVDet (BL) & \textbf{1.74} & \textbf{1.00} & 125.91M &44.1\\ 
			Teacher model & 3.77  & 3.33 & 135.39M  &\textbf{49.4}\\
			RadOcc (ours)  & 2.64  & \textbf{1.00} & \textbf{125.91M} &46.1\\
			\hline
		\end{tabular}
	}
 \caption{\textbf{Complexity analysis.} The analysis of training and inference time (s/sample) and model parameters of BEVDet (baseline), teacher model and our RadOcc.}
 \label{supp:tab1}
\end{table}

\noindent\textbf{Ablation on distillation weights.}
In order to explore the effects of different distillation losses, we conduct several experiments of various combinations of loss weights in the Occ3D validation set with the ResNet-50 backbone as the image-view encoder. The results are shown in Table~\ref{tab:ablation_weight}.
It indicates that every loss term plays an important role in the learning process, and we also found that the $\lambda_{rdc}=100.0$, $\lambda_{sad}=10.0$, $\lambda_{KL}=10.0$ perform best.

\begin{table}[h]
	\footnotesize
	\centering
	\begin{tabular}{c|c|c|cc}
		\toprule
		$\lambda_{rdc}$ & $\lambda_{sad}$ & $\lambda_{KL}$ & mIoU \\\hline
		\midrule
         1.0 & 1.0 & 1.0 & 37.21 \\
         10.0 & 1.0 & 1.0 & 37.43 \\
         100.0 & 1.0 & 1.0 & 37.60 \\
         100.0 & 10.0 & 1.0 & 38.11\\
         100.0 & 10.0 & 10.0 & \textbf{38.53} \\
         1000.0 & 10.0 & 10.0 & 38.31 \\
		\bottomrule
	\end{tabular}
    \caption{\textbf{Ablation for distillation loss weights.} We ablate different weight combinations of the distillation losses.
	}
	\label{tab:ablation_weight}
\end{table}

\noindent\textbf{Ablation on the teacher model.}
We do not constrain the model architecture of the teacher model.
But for distillation, we need a teacher model to achieve at least a higher performance than the baseline student model. 
Here we compare the different teacher models in the occupancy prediction task, including the LiDAR-based models and the multi-modal models.
We use SECOND~\cite{yan2018second} as the backbone and SECOND-FPN~\cite{yan2018second} as the neck to extract the sparse 3D point cloud features.
After that, the 3DCNN is utilized to further aggregate the volume features. 
Since we found the different learning rate schedulers could have effects when training the teacher model, we also ablate the different schedulers.
The results are listed in Table~\ref{tab:abl_teacher}.
From the results, we can easily find that the LiDAR-only-based methods obtained much lower performance than multi-modal methods, despite leveraging the more dense point cloud merged from multi-sweeps.

\begin{table}[t]
	\footnotesize
	\centering
	\begin{tabular}{l|c|cc}
		\toprule
		Method & Scheduler & mIoU \\\hline
		\midrule
	    LiDAR w/ Single Frame &  cyclic &  40.20 \\
        LiDAR w/ Single Frame & linear & 42.16 \\
        LiDAR w/ Multi-sweeps & cyclic & 43.41 \\
        LiDAR w/ Multi-sweeps & linear & 45.07 \\
        LiDAR w/ Multi-sweeps* & linear & 46.01 \\
        w/ Camera* & linear & \textbf{49.38 } \\
		\bottomrule
	\end{tabular}
    \caption{\textbf{Design analysis of teacher model.} We have tried different teacher models on the occupancy prediction task. Since the performances of LiDAR-only-based methods are not enough for knowledge distillation, we choose the multi-modal teacher model in this work. * means we applied the BEV augmentation during training.
	}
	\label{tab:abl_teacher}
\end{table}%



\section{More Visualization Results}
We provide more visual comparisons of RadOcc with our baseline model on the dense 3D occupancy prediction task in Figure~\ref{fig:comparision} and sparse 3D occupancy prediction task, \textit{i.e.} LiDAR semantic segmentation task in Figure~\ref{fig:comparision_lidarseg}.

As can be seen from Figure~\ref{fig:comparision}, our proposed method performs better on moving objects (\textit{e.g.}, car, truck) and small objects (\textit{e.g.} pedestrian) owing to the proposed distillation objective. 
And our method could obtain more clear and regular static structures.
However, the baseline method sometimes always predicts incorrect classes and could generate some voxels that are not occupied in fact.
For example, they tend to mix the categories of terrain, vegetation, driveable surface, and manmade voxels (see the first row and third row in Figure~\ref{fig:comparision}) and could obtain some fault flying occupied voxels overhead (see the bottom row in Figure~\ref{fig:comparision}).


For the point cloud semantic segmentation results, we also visualize the predicted semantic points by using the voxels but more sparse than the dense 3D occupancy prediction results.
From Figure~\ref{fig:comparision_lidarseg}, we can easily find that the baseline has a higher error recognizing small objects and region boundaries, while our proposed method recognizes small objects better thanks to the prior of multi-modality.

\end{document}